\documentclass[letterpaper, 10 pt, conference]{ieeeconf}  
\usepackage{graphics} 
\usepackage{epsfig} 
\usepackage{mathptmx} 
\usepackage{times} 
\usepackage{amsmath} 
\usepackage{amssymb}  
\usepackage{array}
\usepackage{siunitx}
\usepackage{subfigure}
\usepackage{threeparttable}
\usepackage{url}
\usepackage{multirow}
\usepackage{booktabs}

\IEEEoverridecommandlockouts                              

\title{\LARGE \bf
ForceFormer: Exploring Social Force and Transformer for Pedestrian Trajectory Prediction
}

\author{Weicheng Zhang$^{1}$, Hao Cheng$^{2}$, Fatema T. Johora$^{3}$ and Monika Sester$^{1}$
\thanks{$^{1}$Weicheng Zhang and Monika Sester are with the Institute of Cartography and Geoinformatics, Leibniz University Hannover, Appelstr. 9, 30167 Hannover, Germany
        {\tt\small weicheng.zhang@stud.uni-hannover.de, Monika.Sester@ikg.uni-hannover.de}}%
\thanks{$^{2}$Hao Cheng is with the Faculty of Geo-Information Science and Earth Observation, University of Twente, 7500 AE Enschede, The Netherlands
        {\tt\small h.cheng-2@utwente.nl}}%
\thanks{$^{2}$Fatema T. Johora is with the Department of Informatics, Clausthal University of Technology,
Julius-Albert-Str. 4, 38678 Clausthal-Zellerfeld, Germany
        {\tt\small fatema.tuj.johora@tu-clausthal.de}}%
}

\begin{document}

\maketitle
\thispagestyle{empty}
\pagestyle{empty}

\begin{abstract}
Predicting trajectories of pedestrians based on goal information in highly interactive scenes is a crucial step toward Intelligent Transportation Systems and Autonomous Driving. 
The challenges of this task come from two key sources: (1) complex social interactions in high pedestrian density scenarios and (2) limited utilization of goal information to effectively associate with past motion information. 
To address these difficulties, we integrate social forces into a Transformer-based stochastic generative model backbone and propose a new goal-based trajectory predictor called \textit{ForceFormer}. 
Differentiating from most prior works that simply use the destination position as an input feature, we leverage the driving force from the destination to efficiently simulate the guidance of a target on a pedestrian. 
Additionally, repulsive forces are used as another input feature to describe the avoidance action among neighboring pedestrians. 
Extensive experiments show that our proposed method achieves on-par performance measured by distance errors with the state-of-the-art models but evidently decreases collisions, especially in dense pedestrian scenarios on widely used pedestrian datasets.
\end{abstract}

\section{INTRODUCTION}
Accurate and plausible trajectory prediction in crowd scenarios for pedestrians plays a fundamental role in different applications, such as mobile robot navigation~\cite{luo2018porca}, Intelligent Transportation Systems and Intelligent Vehicles~\cite{ballan2016knowledge}, and shared space safety~\cite{li2021autonomous}. 
Unlike vehicle movement governed by traffic rules, such as lane geometry, traffic lights and the headway direction, pedestrians may stop or turn at any time and interact more with neighbors, making their behavior highly stochastic. 

In order to model pedestrian behavior, a variety of different methods have been applied. 
In rule-based models, the interactions among pedestrians, namely social interactions, are described as forces \cite{helbing1995social}. 
In data-driven models, attention mechanisms~\cite{vaswani2017attention} and graph convolutional networks~\cite{huang2019stgat} are widely used to extract social interactions and obtain excellent results using supervised learning \cite{alahi2016social,cheng2021amenet,yuan2021agentformer}.
Goal information can also reduce the uncertainties of pedestrians' behavior. 
However, each of these methods has its own drawbacks. 
Although the rule-based models can simulate a certain degree of pedestrian behavior, they are relatively less robust and resilient in the face of complex scenarios. 
Data-driven models often achieve better performance but are data-dependent and less interpretable~\cite{cheng2021trajectory}. In goal-based models, goal information is often directly applied as an input or as the offset of the current position to the goal~\cite{mangalam2020not,dendorfer2020goal,chiara2022goal}. They are sub-optimal because the association between the goal and the current position is not established.  

To this end, we propose a novel goal-based pedestrian trajectory prediction framework called \textit{ForceFormer}.
It takes as input not only the sequential motion information but also forces to train a Transformer-based backbone. 
Unlike the previous models that directly use last position information as an input feature parallel to other features describing motion dynamics, we apply the goal information to derive social forces so that the changes in velocity, position, and direction are better linked to the goal information. 

In addition, we use the generative model AgentFormer~\cite{yuan2021agentformer} as the trajectory prediction backbone, which utilities the Transformer~\cite{vaswani2017attention} network to learn social interactions in the temporal dimensions.  
Simultaneously, to estimate the temporary goal position for computing forces in the inference time, a goal-estimation module~\cite{mangalam2021goals,chiara2022goal} is applied.  
More specifically, history trajectories are concatenated with semantic scene information, and they are fed into a U-Net~\cite{ronneberger2015u} structure to predict the potential goals. 
With this goal-estimation module, we can obtain reliable goal information and as well naturally take into account the constraints of environmental factors.

In summary, our major contributions are as follows: 
\begin{itemize}
  \item We propose a goal-based trajectory prediction framework \textbf{ForceFormer}. It imports more interpretable features, i.\,e., social forces, into a data-driven model to learn stochastic pedestrian behavior. 
  \item Different variants of ForceFormer making use of goal information are studied, and we find two effective. Namely, a) \textbf{ForceFormer-Re} applies goal positions to derive repulsive forces, reinforcing the interactive information between the ego pedestrian and neighbors and decreasing the possibility of collisions; b) \textbf{ForceFormer-Dr} applies goals to derive driving force, enhancing destination guidance to predict the ego pedestrian's future trajectory.
  \item Extensive empirical studies are carried out on the widely used ETH/UCY \cite{pellegrini2009you,lerner2007crowds} pedestrian datasets. The experimental results show that ForceFormer performs on par with the state-of-the-art models measured by standard distance errors but it evidently decreases collisions, especially in dense pedestrian scenarios.
\end{itemize}

\section{RELATED WORK}
This section briefly reviews the works in sequence modeling, social interaction modeling, and goal-based models.

\vspace{6pt}
\noindent\textbf{Sequence Modeling.} 
Essentially, motion trajectory is composed of positional information on time series. 
Therefore, converting trajectory prediction to sequence-to-sequence modeling is one of the most common approaches. 
In previous works, thanks to their powerful gating functions, Long Short-Term Memories (LSTMs)~\cite{hochreiter1997long} have been widely applied to many pedestrian trajectory prediction tasks and achieved excellent results, especially in the temporal dimension \cite{alahi2016social,huang2019stgat,zhang2019sr,xu2018encoding,cheng2021amenet}. 
In recent years, with the great achievements of the Transformer~\cite{vaswani2017attention} network in the domain of Natural Language Process (NLP) \cite{devlin2018bert,lan2019albert}, Transformer-based models are also applied to trajectory forecasting. 
In contrast to LSTMs, Transformer networks have a better capability of modeling temporal dependencies in long sequences based on the self-attention mechanism~\cite{giuliari2021transformer,yu2020spatio,yuan2021agentformer}. 
In addition to the previously adopted deterministic approaches like LSTMs, an increasing number of deep generative models, such as conditional variational autoencoders (CVAEs) \cite{kingma2013auto, sohn2015learning} and generative adversarial networks (GANs) \cite{goodfellow2020generative} are applied to trajectory forecasting.
Rather than producing one single prediction, generative models learn the potential future trajectories as a distribution and generate multiple possible predictions from latent space. 
For example, Social GAN and Sophie \cite{gupta2018social,sadeghian2019sophie} are proposed for pedestrian multi-path trajectory prediction via jointly training a generator and a discriminator.
Compared to GANs, CVAE models predict multiple plausible trajectories conditioned on the past trajectories and acquire better performance in recent works \cite{ivanovic2019trajectron,tang2019multiple,salzmann2020trajectron++,cheng2021amenet, Chen_2021_ICCV}.

\vspace{6pt}
\noindent\textbf{Social Interaction Modeling.} Besides modeling individual trajectory sequences, establishing the influence of pedestrians on each other or from the environment has been a critical issue in pedestrian trajectory prediction.
As groundbreaking work, Helbing et al. \cite{helbing1995social} leverage dynamic social forces to imitate the influence of the surroundings on pedestrians, e.\,g., a repulsive force for collision avoidance and an attractive force for social connection.
The social force model has been effectively applied in various fields like robotics \cite{ferrer2013robot} and crowd analysis \cite{mehran2009abnormal,johora2022generalizability}. 
Another pioneering data-driven work is Social LSTM \cite{alahi2016social}. It proposes a new structure, called the social pooling layer, to aggregate the interaction information from neighbors.
With the development of graph neural networks (GNNs) \cite{kipf2016semi}, more recent works of deterministic models like \cite{yu2020spatio,kosaraju2019social} resort to modeling a crowd as a graph and combining GNNs with attention mechanisms to learn spatial interactions. 
Other approaches like \cite{salzmann2020trajectron++,huang2019stgat} first encode features over social dimension at each independent time step.
Then, these social features are fed into another temporal sequence model to summarize the social relations over time. 
Unlike these methods above, we import social forces at each time step to a Transformer-based backbone, facilitating the learning of social interactions among pedestrians.

\vspace{6pt}
\noindent\textbf{Goal-based Model.} Recently, goal-based models have become an effective way to improve prediction performance \cite{mangalam2020not,Zhao_2021_ICCV}.
Diverse goal information can provide more predictive possibilities to deterministic model~\cite{zhao2020tnt,chiara2022goal}. Moreover, pedestrians are motivated by their destinations. Therefore, high uncertainty behavior can be limited through the goal information \cite{mangalam2021goals}.
In contrast to those models that directly use the goal information as an input feature, we use the goal information to calculate the social forces for each pedestrian \cite{helbing1995social}, and the resulting forces are used as input to our prediction module.
However, the goal information is not accessible in inference time.
To circumvent this issue, we utilize a goal-estimation \cite{mangalam2021goals} module for estimating the goals in the inference time.

\section{METHODOLOGY}
\subsection{Problem formulation}
In the context of pedestrian trajectory prediction problems, a complete trajectory of a pedestrian can be divided into two parts, the observed and the future trajectories. 
The observed trajectory at time steps $t \leq 0$ is denoted as $X = (X^{-H}, X^{-H+1}, . . . X^0)$, which in total includes $ H+1 $ observed time steps; 
While, the future trajectory at time steps $t > 0$ is denoted as $Y = (Y^1, Y^2, . . ., Y^T)$ over $ T $ future time steps. 
Similar to \cite{yuan2021agentformer}, we use the $x$- and $y$-coordinate and the velocity sequence in the 2D coordinate system to parameterize trajectories. 
In addition, the joint social sequences of all $ N $ pedestrians in the same scene at the same time step $t$ are denoted as $X^t = (x^t_1, x^t_2, . . ., x^t_N)$ for the observation and $Y^t = (y^t_1, y^t_2, . . ., y^t_N)$ for the future trajectories. 
In our proposed generative model $p_{\theta}(Y|X, G, F)$, where $\theta$ are the model parameters, the task is to forecast future trajectories $Y$ depending on not only observed trajectories $X$ but also goal information $G$ and social forces $F$. 

Following \cite{chiara2022goal}, we use both the position of the last time step and the differences between every single position and the goal position to parameterize the goal information. 
It should be noted that we use the ground truth of the last position $Y^T$ to derive the goal representation in the training phase, while we use the estimation from the goal-estimation module in~Sec.~\ref{framework} to derive the goal representation in the test phase. 
Additionally, two kinds of forces, i.\,e., driving force $F_\text{Dr}$ and repulsive force $F_\text{Re}$, are calculated for each agent at every time step. They are also represented as sequences. 
\begin{equation} \label{two kinds of forces}
F = \left\{ \begin{array}{ll}
F_\text{Dr} = ({f_\text{Dr}}_1^{-H},...,{f_\text{Dr}}_N^{-H},...,{f_\text{Dr}}_1^{T},...,{f_\text{Dr}}_N^{T}),\\
F_\text{Re} = ({f_\text{Re}}_1^{-H},...,{f_\text{Re}}_N^{-H},...,{f_\text{Re}}_1^{T},...,{f_\text{Re}}_N^{T}).
\end{array} \right.
\end{equation}

\begin{figure}[ht]
\centering
\includegraphics[trim=0.2in 0.1in 0.3in 0.0in, clip=true, width=\linewidth]{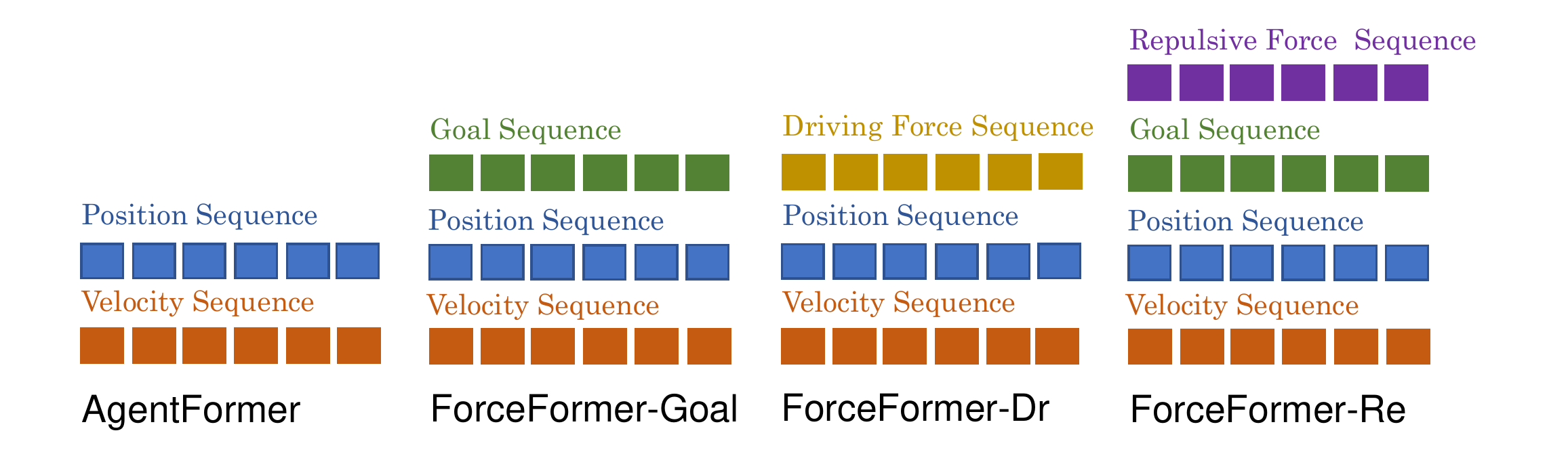}
\caption{Inputs of AgentFormer~\cite{yuan2021agentformer} and the proposed ForceFormer-Goal, ForceFormer-Dr, and ForceFormer-Re.}
\label{fig:Inputscomparison}
\end{figure}

In order to explore different ways of incorporating the goal information, on the basis of the baseline model AgentFormer~\cite{yuan2021agentformer} that takes velocity and position sequences as input, we propose three variants of the additional goal information. 
As denoted in Figure~\ref{fig:Inputscomparison}, \textbf{ForceFormer-Goal} directly adds the additional goal sequence to the input.
Alternatively, \textbf{ForceFormer-Dr} uses $ F_\text{Dr} $ as the additional conditional information.
\textbf{ForceFormer-Re} uses both the goal sequence and the repulsive force $ F_\text{Re}$ sequence as the additional input. 

\subsection{The Proposed Framework}
\label{framework}
\begin{figure*}[ht]
\centering
\includegraphics[width=1\textwidth]{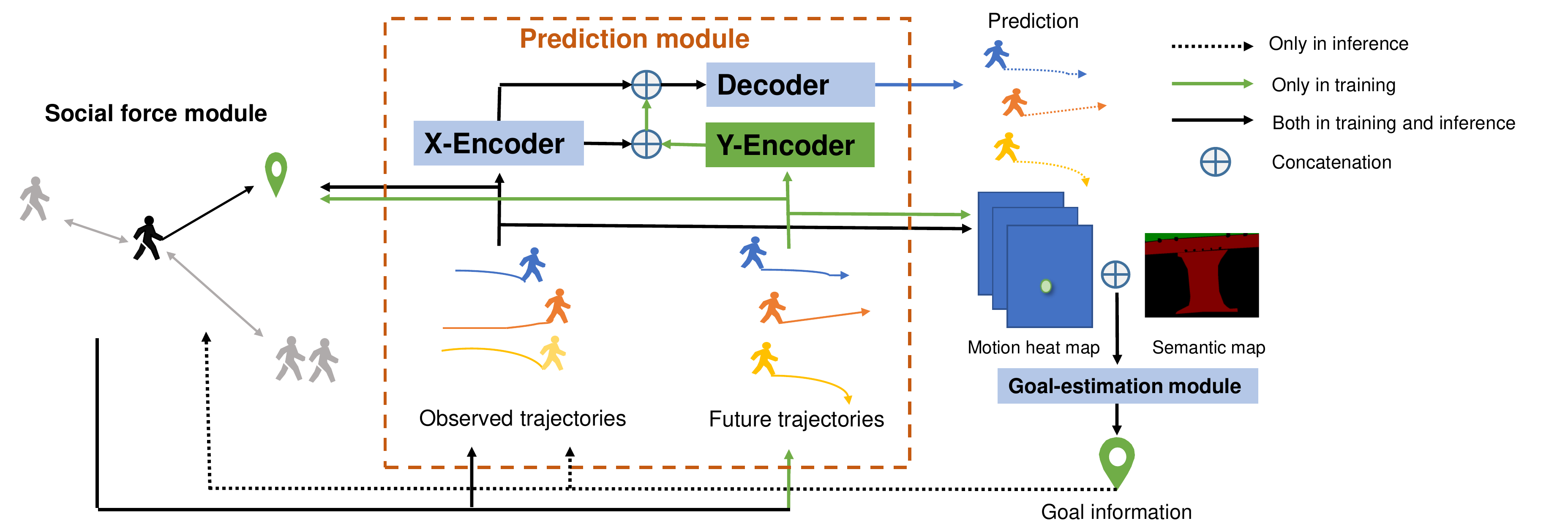}
\caption{An overview of the proposed framework {ForceFormer}.}
\label{fig:Social-AgentFormer}
\end{figure*}

Figure~\ref{fig:Social-AgentFormer} depicts the overview of our proposed framework ForceFormer. It mainly consists of three modules: AgentFormer (X-Encoder, Y-Encoder, and Decoder) as the backbone, Goal-estimation module, and Social force model.

In the training process, the goal-estimation module and AgentFormer are trained separately. 
The goal information is supplied from ground truth $ Y^T $, which is used for training the goal-estimation module and the calculation of social forces.
The repulsive force and driving force are calculated by the position information, velocity information, and goal information.
However, the ground truth $ Y^T $ is unavailable in the test phase.
Hence, during the inference process, we sample $K$ goal candidates for every trajectory from the goal-estimation module.
Following the previous goal-conditioned trajectory prediction models \cite{Zhao_2021_ICCV,mangalam2020not,mangalam2021goals}, we evaluate all potential $K$ goals against the ground truth and choose the one with the smallest $L2$ error as the estimated goal position in the test phase.

\paragraph{AgentFormer} 
The backbone prediction model is a CAVE-based model and establishes spatial and temporal relations using attention mechanisms. 
Based on the conditions of observed trajectory $X$, goal information $G$, and social forces $F$, the future trajectory distribution is modeled as $p_{\theta}(Y|X,G,F)$. 
The future trajectory distribution can be rewritten as
\begin{equation} \label{future distribution}
p_{\theta}(Y|X,G,F) = \int p_{\theta}(Y|Z,X,G,F)p_{\theta}(Z|X,G,F)dZ,
\end{equation}
where $p_{\theta}(Z|X,G,F)$ is the conditional Gaussian prior, which is learned by X-Encoder.
$ p_{\theta}(Y|Z,X,G,F)$ is the conditional likelihood. 
Eq.~\eqref{future distribution} proposes a set of latent variables $Z = ( z^{(1)},...,z^{(K)} )$, reflecting the latent intent of pedestrian $ n $ to account for stochasticity and multi-modality in the pedestrian's future behavior. 

The negative evidence lower bound $\mathcal{L}_{elbo}$ is used to address the intractable posterior $p_{\theta}(Z|Y,X,G,F)$. Concretely, the CVAE-based model is optimized using the loss function
\begin{equation} \label{CVAE lower bound}
\begin{aligned}
\mathcal{L}_{elbo} = -\mathbb{E}_{q_{\phi}(Z|Y,X,G,F)}[\log p_{\theta}(Y|Z,X,G,F)] \\+ KL(q_{\phi}(Z|Y,X,G,F)||p_{\theta}(Z|X,G,F)),
\end{aligned}
\end{equation}
where $q_{\phi}(Z|Y,X,G,F)$ is the approximate posterior distribution parameterized by $\phi$, which is learned by Y-Encoder.
The first term in the above equation can be considered as the expected predicted probability of the future trajectory $ p_{\theta}(Y|Z,X,G,F) $. 
The second term $KL(q_{\phi}(Z|Y,X,G,F)||p_{\theta}(Z|X,G,F))$ denotes the distribution difference between the prior and the approximate posterior, which both tend to be a standard normal distribution.

\paragraph{Social Forces}
A modified social force model \cite{helbing1995social} that contains both driving and repulsive forces is applied in this work. 
The \textbf{driving force}, denoted by Eq.~\eqref{driving force},  describes the attractive effect related to the destination (goal position).
\begin{equation} \label{driving force}
\Vec{F}_\alpha^0 = \frac{1}{\tau_\alpha}(v_\alpha^0\Vec{e}_\alpha - \Vec{v}_\alpha).
\end{equation}
The value of the driving force depends on the deviation of the current velocity $\Vec{v}_\alpha(t)$ from the desired velocity $v_\alpha^0(t) = v_\alpha^0\Vec{e}_\alpha$. 
Velocity $v_\alpha^0(t) = v_\alpha^0\Vec{e}_\alpha$ denotes that if a pedestrian is not disturbed, she will walk at the desired speed $v_\alpha^0$ along the desired direction $\Vec{e}_\alpha$ pointing to the destination. 
The relaxation time $\tau_\alpha$ is a parameter that represents the expected time removing this deviation.

To avoid collisions, pedestrians maintain a proper distance from other strangers.
The \textbf{repulsive force}, denoted by Eq.~\ref{repulsive effect}, describes the avoidance phenomenon between the ego pedestrian $\alpha$ and other pedestrian $\beta$, 
\begin{equation} \label{repulsive effect}
\Vec{f}_{\alpha \beta}(\Vec{r}_{\alpha \beta}) = -\nabla_{\Vec{r}_{\alpha \beta}}V_{\alpha \beta}[b(\Vec{r}_{\alpha \beta})].
\end{equation}
The repulsive potential $V_{\alpha \beta}(b)$ is a monotonic decreasing function related to $ b $, which represents the semi-minor axis of an ellipse. 
Through $ b $, the equipotential lines keep the form of an ellipse that is pointed to the direction of motion.
\begin{equation} \label{semi-minor axis}
2b = \sqrt{(\left\|\Vec{r}_{\alpha \beta}\right\| + \left\|\Vec{r}_{\alpha \beta} - v_\beta\Delta t \Vec{e}_\beta\right\|)^2 - (v_\beta\Delta t)^2}.
\end{equation}

In addition to distance, the influence of viewpoint also needs to be considered when calculating repulsive forces.
Thus a parameter $ w $ is introduced.
\begin{equation} \label{feld of view}
w(\Vec{e},\Vec{f}) = \left\{ \begin{array}{ll}
1 & \textrm{if $ \Vec{e} \cdot \Vec{f} \geq \left\| \Vec{f} \right\|cos\epsilon $},\\
c & \textrm{otherwise},
\end{array} \right.
\end{equation}
where the effective angle of sight is $2\epsilon$, and $c$ is a constant factor.
Hence, the repulsive force, after taking the perspective factor into account, is constrained as $ \Vec{F}_{\alpha \beta}= w\Vec{f}_{\alpha \beta} $.

Since the repulsive force is based on the premise that two pedestrians are strangers, they want to keep their distance from each other and avoid collisions. However, in reality,
many pedestrians travel in pairs, such as classmates, relatives, and friends, who share a common destination. 
Therefore, it is not logical to consider their repulsive force within a group, which could cause significant errors in the experimental results, especially in high pedestrian density scenes. 
So we adapt the DBSCAN method, which is a density-based spatial clustering of applications with noise~\cite{ester1996density,schubert2017dbscan} for every time step. 
Through the clustering, the candidates of group members are detected~\cite{cheng2019pedestrian}. 
Namely, if two pedestrians are in the same cluster for more than $ \sigma $ time steps in the observed frames, they will be judged to be in the same group.
Then the intra-group repulsive forces are eliminated.

\paragraph{Goal-Estimation Module}
The goal-estimation module is employed to provide the goal information for the AgentFormer prediction module and the calculation of social forces in the inference phase. 
We adopt the goal module proposed in \cite{mangalam2021goals,chiara2022goal} for this purpose. 
First, the past trajectories in a heat map form concatenates with the semantic map information. 
The semantic map is adopted from Chiara et al.~\cite{chiara2022goal}, which is extracted from a bird's-eye view of the RGB scene image in ETH/UCY datasets using a pre-trained segmentation network \cite{mangalam2021goals}.
More specifically, through the semantic map, the constraints of environments, for instance, pavement, terrain, and building, can be naturally considered. 
The segmentation results in a tensor form $ S \in \mathbb{R}^{W\times L \times C} $ containing $ C $ classes.
$H$ and $L$ are the height and width of the input image. 
The past trajectory$ \left\{ x_n^{-H}, x_n^{-H+1},...,x_n^0 \right\} $ of agent $n$ is mapped to the heat map $ M \in \mathbb{R}^{W\times L \times (H+1)}  $. 
Then, the heat map tensor of the past trajectory is concatenated with the semantic map $ S $ along the channel dimension, generating tensor $ M_s \in \mathbb{R}^{W\times L \times (C+H+1)}  $ as the input tensor for goal estimation.
Finally, the concatenated information is fed to a U-Net~\cite{ronneberger2015u} model to generate a probability map of future positions.

\section{EXPERIMENTS}
\subsection{Dataset}
The proposed framework is evaluated on ETH \cite{pellegrini2009you} and UCY \cite{lerner2007crowds}, which have been widely used as the benchmark for pedestrian trajectory prediction.
The datasets contain five different subsets as listed in Table~\ref{pedestrian density}.
A valid trajectory denotes a single pedestrian's track information in 20 consecutive frames captured at 2.5\,Hz. 
These twenty-time steps are divided into two parts -- the first eight time steps (3.2\,s) are observed trajectories $ X $, based on which twelve future time steps (4.8\,s) as future trajectories $ Y $ are predicted.
The position of the goal is located at the twentieth time step.
It can be seen that the density of pedestrians varies across the subsets.
The density in a scene largely influences the prediction results, especially in this work, because the calculation of social forces is closely related to crowd density. 

\begin{table}[h!]
\begin{centering}
\caption{The number of frames and valid trajectories in the ETH/UCY \cite{pellegrini2009you,lerner2007crowds} datasets.}
\begin{tabular}{ m{1.8cm} | m{0.8cm}  m{0.8cm}  m{0.8cm}  m{0.8cm}  m{0.8cm} } 
\toprule
  & ETH & Hotel & Univ & zara01 & zara02 \\ 
\midrule
Frame & 1142 & 1788 & 947 & 883 & 1033\\ 
\hline
Valid trajectory & 364 & 1197 & 24334 & 2356 & 5910 \\
\bottomrule
\end{tabular}
\label{pedestrian density} 
\end{centering}
\end{table}

\subsection{State-of-the-art models and baseline}
We compare our proposed method, ForceFormer, with the following models.
AgentFormer~\cite{yuan2021agentformer} is the baseline model without using any goal information.
Sophoie~\cite{sadeghian2019sophie} proposes a GAN-based model that combines trajectory information with context information.
Trajectron++ \cite{salzmann2020trajectron++} is a CVAE-based model that maintains top performance on the ETH/UCY benchmark.
STAR~\cite{yu2020spatio} proposes a Temporal Transformer and a Spatial Transformer to model spatial-temporal information for pedestrian trajectory prediction.
Moreover, ForceFormer is compared with a bunch of goal-based models. Namely, PECNet \cite{mangalam2020not} is a goal-conditioning model for short-term trajectory prediction. Goal-GAN \cite{dendorfer2020goal} integrates goal information in a GAN-based model for trajectory prediction. Heading \cite{Zhao_2021_ICCV} proposes a goal retrieval module that provides goal information for trajectory prediction. Y-net \cite{mangalam2021goals} combines scene information with goals and waypoints for trajectory prediction. Goal-SAR \cite{chiara2022goal} proposes an attention-based recurrent network combined with the same goal-estimation module as ForceFormer.

 \subsection{Evaluation Metrics and Protocol}
Three metrics are used to evaluate the proposed model. 
First, two standard error metrics are applied to measure the trajectory prediction performance of ForceFormer and compare it fairly with the previous models.
These two distance errors are average displacement error \textbf{$ ADE_K $} and final displacement error \textbf{$ FDE_K $} of $ K $ trajectory samples of each agent compared to the corresponding ground truth.
\begin{equation} 
ADE_K = \frac{1}{T} \text{min}_{k=1}^K\sum_{t=1}^T\left\|\hat{y}^{t,(k)}_n - y^t_n \right\|^2,
\end{equation}
\begin{equation} 
FDE_K = \text{min}_{k=1}^K\left\|\hat{y}^{T,(k)}_n - y^T_n \right\|^2.
\end{equation}

In addition, the number of collisions as another metric is leveraged to verify the social forces applied to ForceFormer.
\begin{equation} 
NC = \sum^N_{m,n=1}\sum^T_{t=1}(\left\|\hat{y}^t_n - \hat{y}^t_m \right\|)< \gamma, m \not= n,
\end{equation}
where $m$ and $n$ are different pedestrians in the same scene at time step $t$. The threshold $ \gamma $ is set for determining whether a collision occurs.
In this paper, $ \gamma = 0.1\,m$.
All the metrics are computed with $K=20$ samples.
The calculation of $ NC $ is based on trajectories with the best $ ADE $.
Following prior works \cite{cheng2022gatraj,yuan2021agentformer,salzmann2020trajectron++,mangalam2020not}, we adopt the leave-one-out strategy for the evaluation.

\subsection{Implementation Details}
For calculating social forces, we adopt \ang{200} as the effective angle $ 2\epsilon $. The factor $ c $ for the field of view is 0.5. The threshold of minimum frames in the same cluster for grouping $\sigma$ is four. 
Also, we consider that the desired direction cannot be calculated when the pedestrian position overlaps with the goal position, so we set the social forces at these positions to zero. For the AgentFormer backbone, we use all the same settings as in Yuan et al.~\cite{yuan2021agentformer}.
But we only train the CVAE model using Adam optimizer \cite{kingma2014adam} for 50 epochs, shorter than the original paper.
For the goal-estimation module, in addition to the same settings as in Chiara et al., \cite{chiara2022goal}, we add a goal-specific MSE loss function $  \mathcal{L}_{\text{MSE}} = \frac{1}{N}\sum_{i=1}^{N} \left\| y_i^T - \hat{y}_i^T \right\|^2$  and a hyper-parameter $\lambda = 1e^6$ to balance the original BCE loss function. 
All our models are trained on Google Colab with a single Tesla P100 GPU.

\subsection{Results}
In table \ref{Quantitative comparisons of all the baselines}, we compare our approaches with current state-of-the-art methods.
First, our proposed methods ForceFormer-Dr and ForceFormer-Re achieve better performance in all the subsets compared to the baseline model AgentFormer, e.\,g., on average, ForceFormer-Dr reduces FDE by $ 26\% $  and ForceFormer-Re reduces ADE by $ 17\% $.
In addition, when comparing to models that also use goal information, our methods perform on par with the previous best method Y-net. 
In particular, when we compare the results on each subset, we can find that our models achieve better performance than Y-net on the other four subsets, except for ETH.
Moreover, when comparing the results in the high-density scenes, i.\,e., on Univ and Zara2, ForceFormer-Dr decreases FDE by 22\% and 23\%, respectively, than Y-net.
The improvements indicate better performance of our method on final position predictions in high-density scenes.

\begin{table}[h!]\scriptsize
\begin{centering}
\caption{Quantitative performances of the state-of-the-art models and our proposed models on the {ETH/UCY} datasets.} 
\begin{threeparttable}
\setlength{\tabcolsep}{0.85mm}
\begin{tabular}{l|cccccc} 
\toprule
\textbf{Method} & \multicolumn{6}{c}{$ ADE_k/FDE_k (m)\, K=20 $ Samples} \\ 
\midrule
Datasets & ETH  & Hotel  & Univ  & Zara1 & Zara2 & Average \\
Sophie \cite{sadeghian2019sophie} & 0.70/1.43  & 0.76/1.67  & 0.54/1.24 & 0.30/0.63  & 0.38/0.78 & 0.54/1.15  \\
STAR \cite{yu2020spatio} & 0.36/0.65 & 0.17/0.36 & 0.31/0.62 & 0.26/0.55 & 0.22/0.46 & 0.26/0.53                 \\
Trajectron++ \cite{salzmann2020trajectron++} & 0.67/1.18 & 0.18/0.28 & 0.30/0.54 & 0.25/0.41 & 0.18/0.32 & 0.32/0.55          \\ 
AgentFormer \cite{yuan2021agentformer} & 0.45/0.75 & 0.14/0.22 & 0.25/0.45 & 0.18/0.30 & 0.14/0.24 & 0.23/0.39      \\
\underline{PECNet} \cite{mangalam2020not} & 0.54/0.87 & 0.18/0.24 & 0.35/0.60 & 0.22/0.39 & 0.17/0.30 & 0.29/0.48        \\
\underline{Goal-GAN} \cite{dendorfer2020goal} & 0.59/1.18 & 0.19/0.35 & 0.60/1.19 & 0.43/0.87 & 0.32/0.65 & 0.43/0.85        \\
\underline{Heading} \cite{Zhao_2021_ICCV} & 0.37/0.65 & 0.11/0.15 & \textbf{0.20}/0.44 & 0.15/0.31 & \textbf{0.12}/0.26 & 0.19/0.36 \\
\underline{Y-net} \cite{mangalam2021goals} & \textbf{0.28}/\textbf{0.33} & 0.10/\textbf{0.14} & 0.24/0.41 & 0.17/0.27 & 0.13/0.22 & \textbf{0.18}/\textbf{0.27} \\                
\underline{Goal-SAR} \cite{chiara2022goal} & \textbf{0.28}/0.39 & 0.12/0.17 & 0.25/0.43 & 0.17/0.26 & 0.15/0.22 & 0.19/0.29 \\
 \hline
 {ForceFormer-Dr} & 0.43/0.58 & 0.12/0.16 & 0.21/\textbf{0.32} & \textbf{0.14}/\textbf{0.20} & \textbf{0.12}/\textbf{0.17} & 0.20/0.29 \\
 {ForceFormer-Re} & 0.36/0.52 & \textbf{0.09}/\textbf{0.14} & 0.21/0.42 & 0.15/0.22 & \textbf{0.12}/0.20 & 0.19/0.30 \\
\bottomrule
\end{tabular}
\begin{tablenotes}
\footnotesize
\item[*] The results of Trajectron++ and Heading are updated according to the implementation issue $ 53 $ \cite{trjectron++.org} and sampling trick \cite{heading.org}. The underlined methods use goal information.
\end{tablenotes}
\end{threeparttable}
\label{Quantitative comparisons of all the baselines}
\end{centering}
\end{table}

\subsection{Ablation study}
The variants of our proposed model making use of the goal information are compared against the FDE values in Table~\ref{AblationFDE} and the number of collisions in Table~\ref{Ablationcollision}. 
First, it can be seen clearly that, compared to the baseline model AgentFormer, all the variants making use of the additional goal information achieve smaller average FDE. 
Except ForceFormer-Goal, ForceFormer-Dr and ForceFormer-Re have evidently smaller numbers of collisions. 
Among the three variants of ForceFormer, ForceFormer-Goal, in general, performs worse than the other models in terms of FDE and the number of collisions across the subsets.
This indicates that directly utilizing the goal sequences may not as effective as the social forces.
With closer observation, we can see that ForceFormer-Dr achieves the smallest FDE in high-density pedestrian scenes like Univ and Zara02. In contrast, ForceFormer-Re has the smallest total collisions, with a 19.8\% reduction compared to the baseline model AgentFormer.

\begin{table}[t]
\begin{centering}
\setlength{\tabcolsep}{0.5mm}
\caption{Performance differences for the methods using goal information measured by FDE.} 
\begin{tabular}{l|ccc|ccccc|c} 
\toprule
 & \multicolumn{8}{c}{$ FDE_k (m) K=20 $ Samples} \\
\hline
{Method} & Goal   & $ F_{Dr} $ &  $ F_{Re} $& ETH  & Hotel  & Univ  & Zara1 & Zara2 & Average \\ 
\midrule
{AgentFormer} \cite{yuan2021agentformer}  & &  &  & 0.75 & 0.22 & 0.45 & 0.30 & 0.24 & 0.39 \\ 
{ForceFormer-Goal} & \checkmark &  &  & 0.55 & 0.17 & 0.49 & 0.30 & 0.28 & 0.36 \\ 
{ForceFormer-Dr}  &  & \checkmark   & & 0.58 & 0.16 & \textbf{0.32} & \textbf{0.20} & \textbf{0.17} & \textbf{0.29} \\
{ForceFormer-Re}&\checkmark  &  & \checkmark & 0.52 & \textbf{0.14} & 0.42 & 0.22 & 0.20 & 0.30 \\
\bottomrule
\end{tabular}
\label{AblationFDE}
\end{centering}
\end{table}

 \begin{table}[ht!]
\begin{centering}
\setlength{\tabcolsep}{0.6mm}
\caption{Performance differences for the methods using goal information measured by collision numbers.} 
\begin{tabular}{l|ccc|ccccc|c} 
\toprule
 & \multicolumn{8}{c}{\textbf{Collision number $ CN_K, K = 20 $ samples}}   \\ 
\hline
  \textbf{Method} & Goal  & $ F_{Dr} $  &  $ F_{Re} $& ETH  & Hotel  & Univ  & Zara1 & Zara2 & Sum \\ 
\midrule
AgentFormer \cite{yuan2021agentformer}  & & & & 0 & 2 & 655 & 4 &  22 & 683 \\
ForceFormer-Goal &\checkmark  &  &  & \textbf{0} & \textbf{0} & 672 & \textbf{3} &  22 & 697 \\ 
ForceFormer-Dr &  & \checkmark &  & \textbf{0} & 1 & 556 & \textbf{3} &  28 & 588\\
ForceFormer-Re  & \checkmark &  & \checkmark & \textbf{0} & 1 & \textbf{529} & 5 &  \textbf{13} & \textbf{548} \\
\bottomrule
\end{tabular}
\label{Ablationcollision}
\end{centering}
\end{table}

\subsection{Qualitative results}
\begin{figure*}[t]
     \centering
     \subfigure{
     \includegraphics[trim=0.0in 0.1in 0.0in 0.0in, clip=true,width=0.3\linewidth]{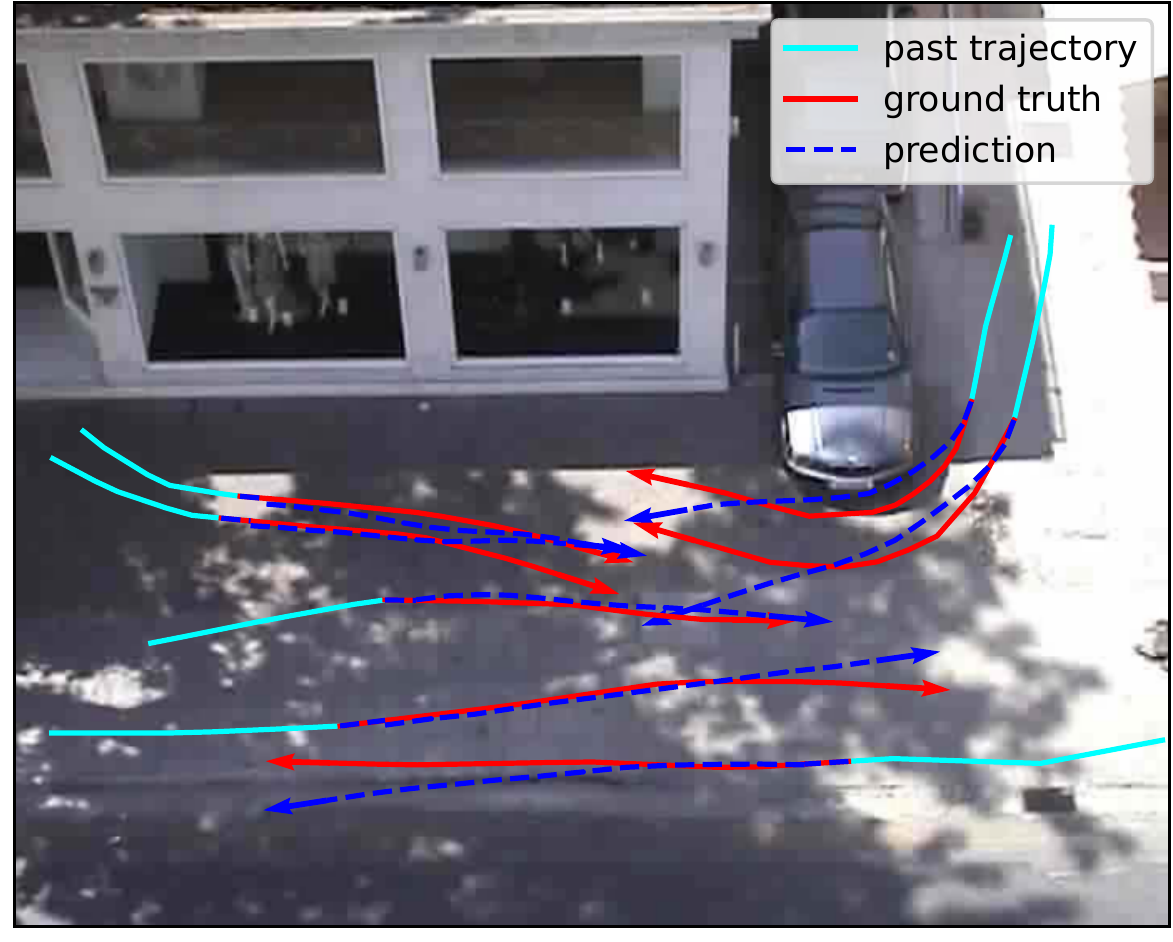}
     }
     \quad
     \subfigure{
         \includegraphics[trim=0.0in 0.1in 0.0in 0.0in, clip=true,width=0.3\linewidth]{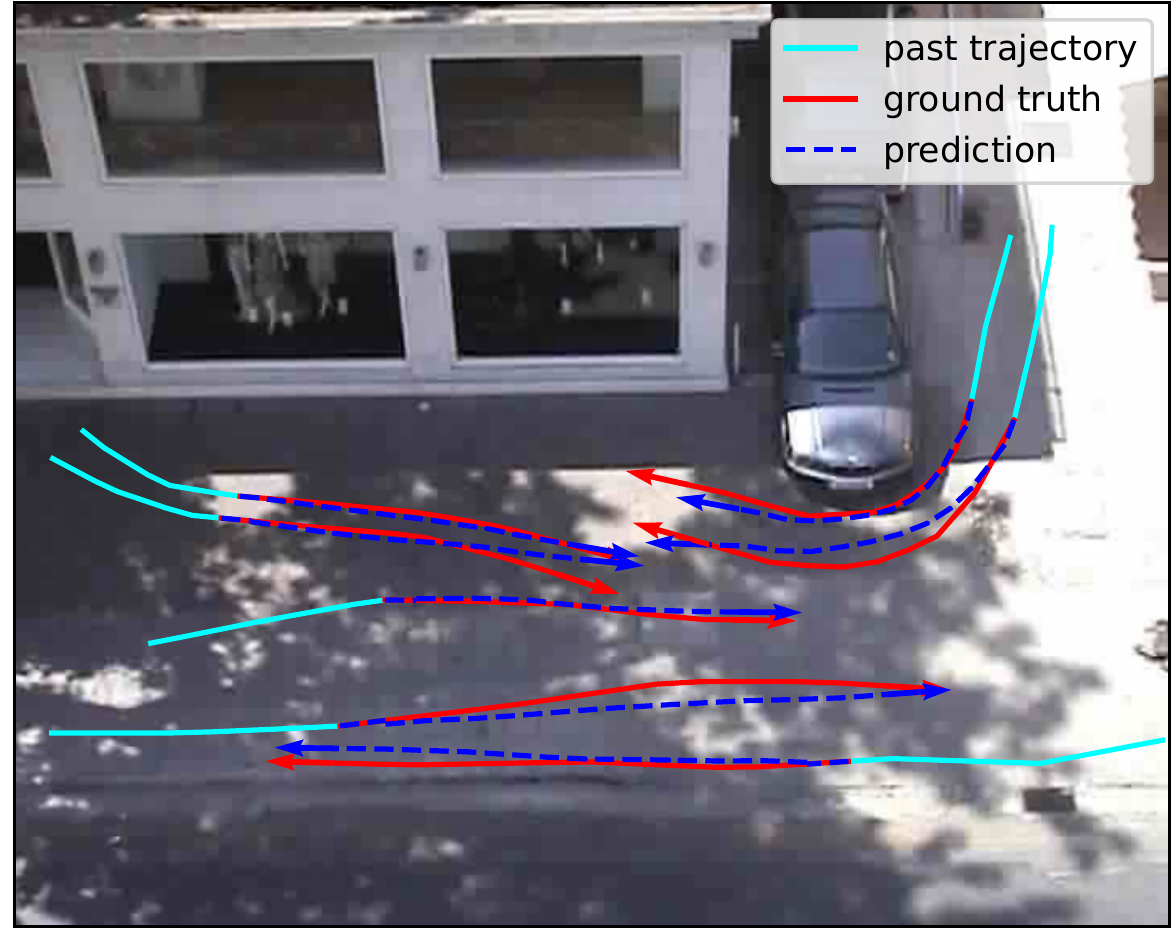}}
     \quad
     \subfigure{
         \includegraphics[trim=0.0in 0.1in 0.0in 0.0in, clip=true,width=0.3\linewidth]{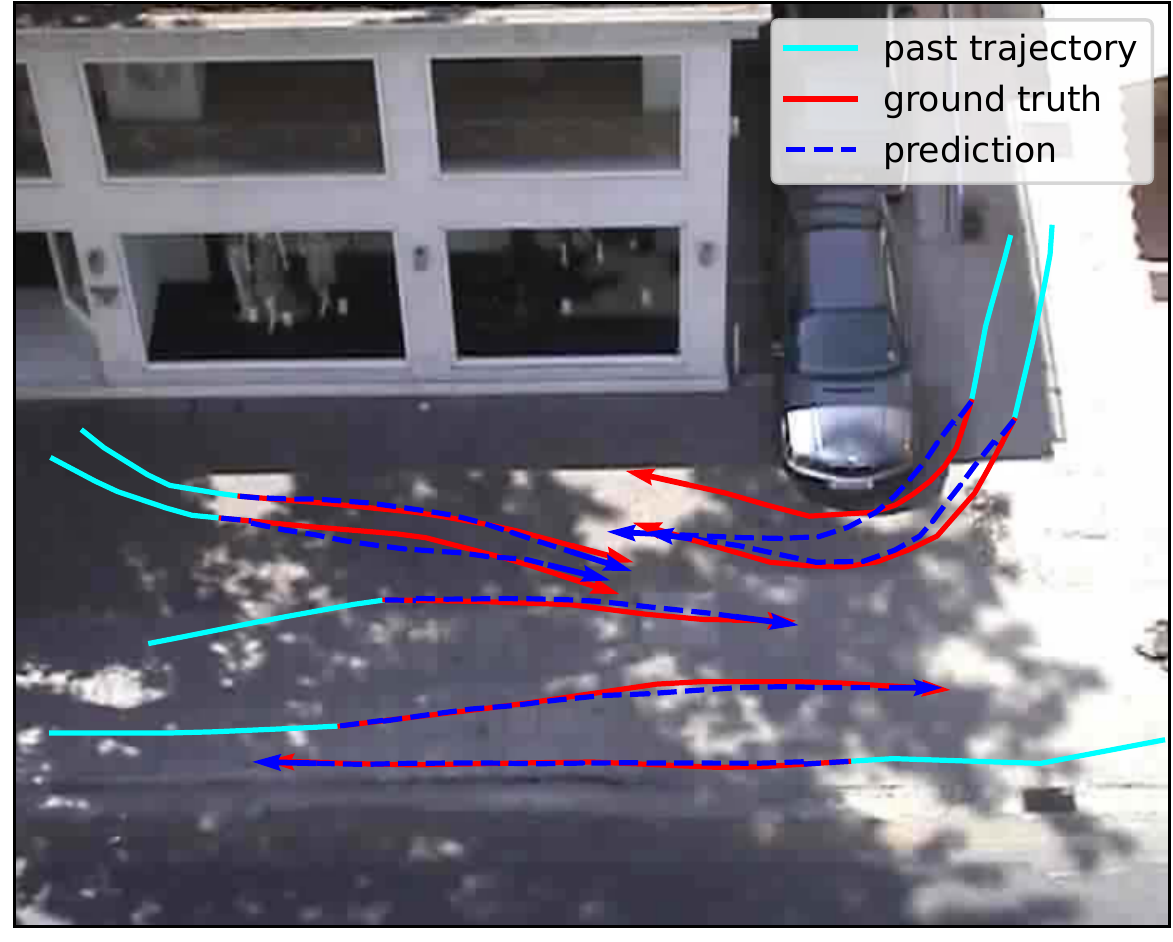}}\

    \subfigure{
     \includegraphics[trim=0in 0.1in 0.0in 0.0in, clip=true,width=0.3\linewidth]{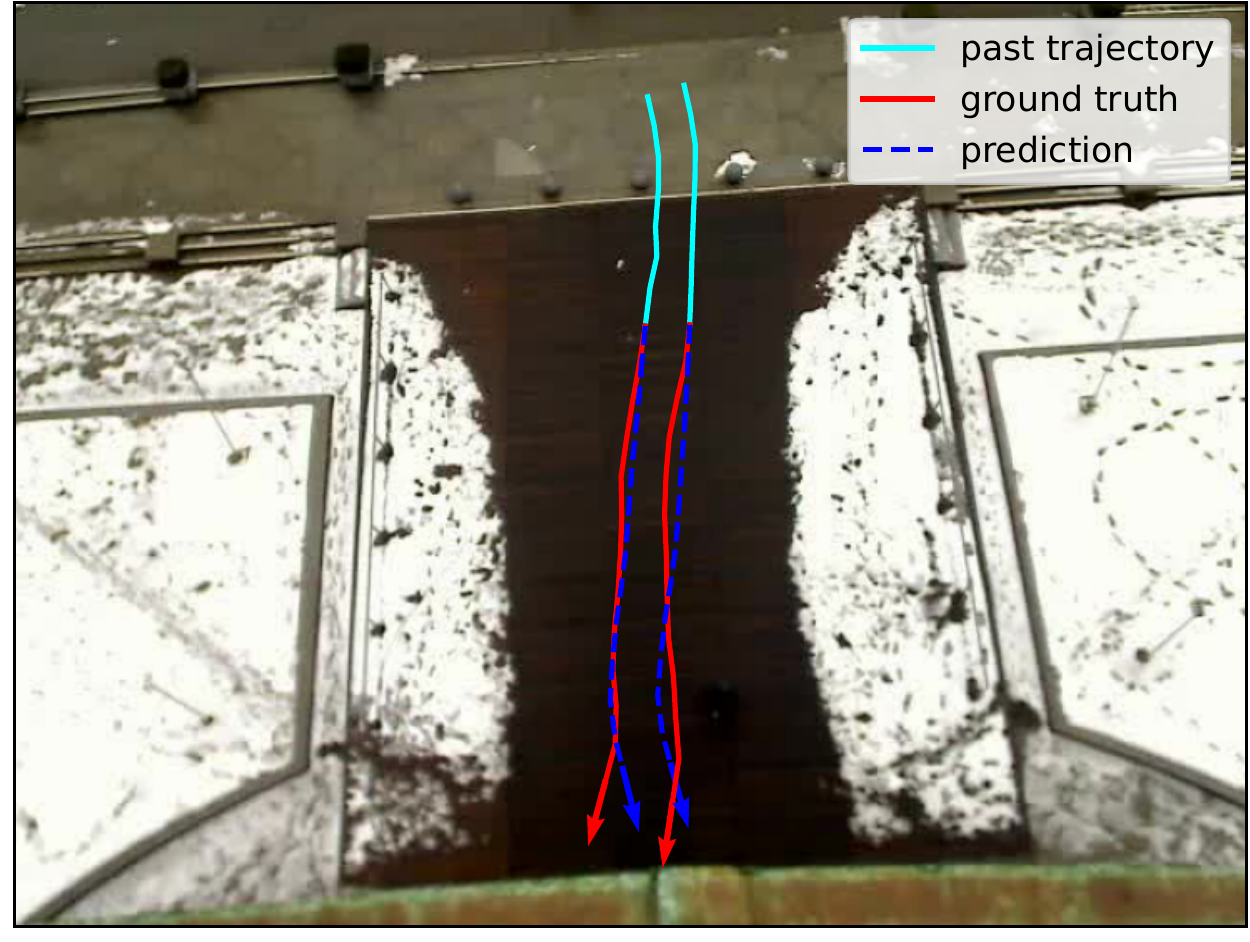} 
     }
     \quad
     \subfigure{
         \includegraphics[trim=0in 0.1in 0.0in 0.0in, clip=true,width=0.3\linewidth]{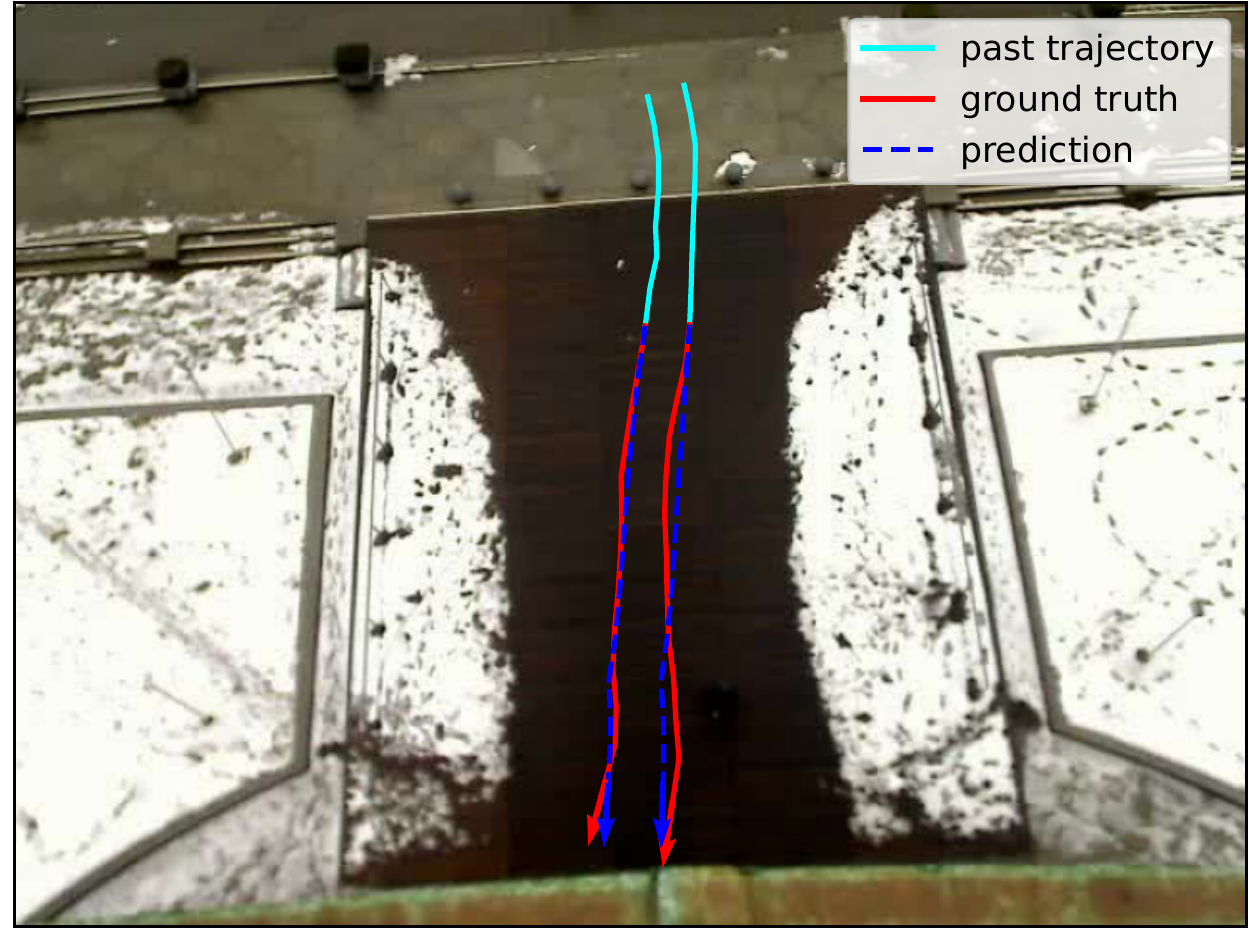}}
     \quad
     \subfigure{
         \includegraphics[trim=0in 0.1in 0.0in 0.0in, clip=true,width=0.3\linewidth]{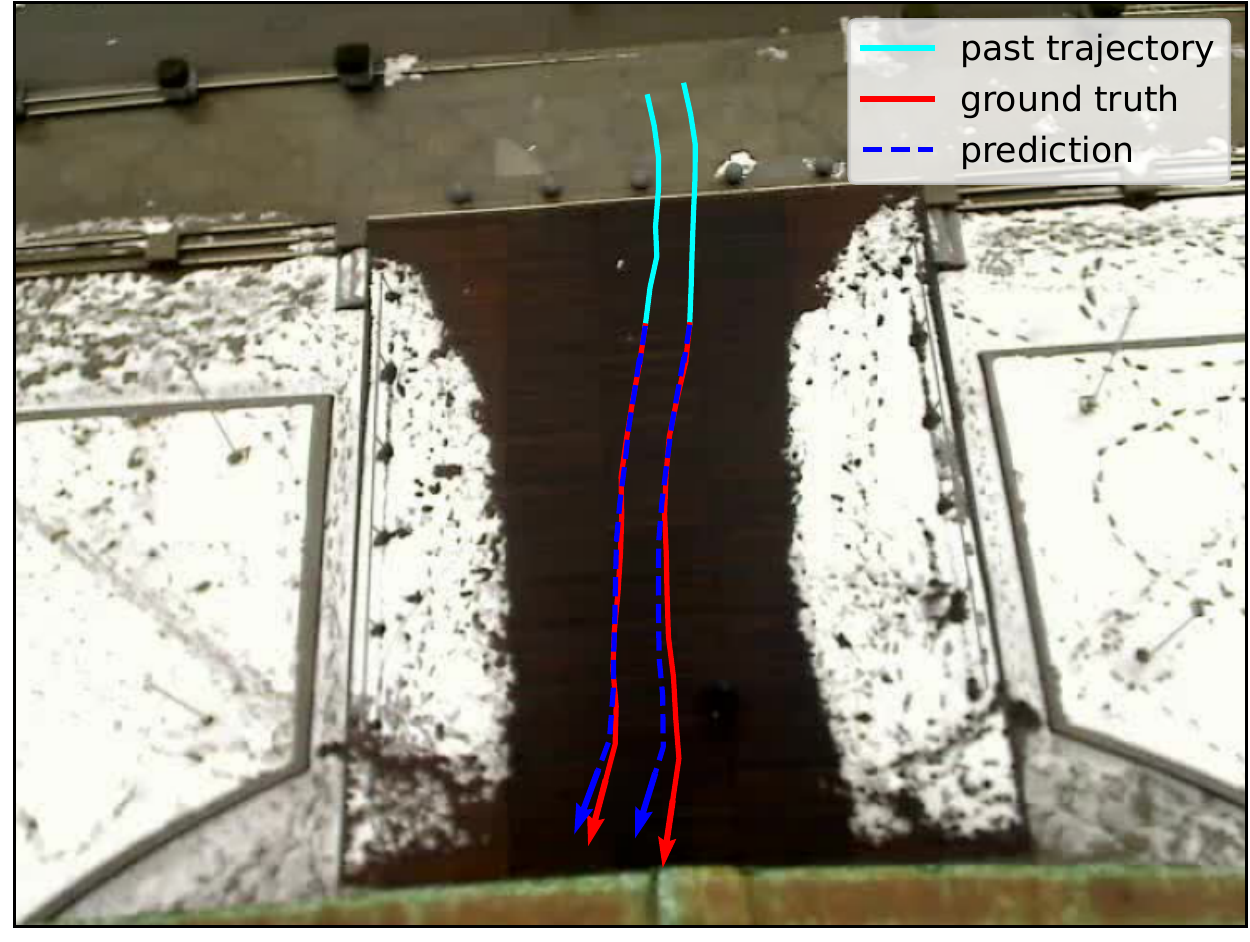}}

        \subfigure{
     \includegraphics[trim=0in 0.1in 0.0in 0.0in, clip=true,width=0.3\linewidth]{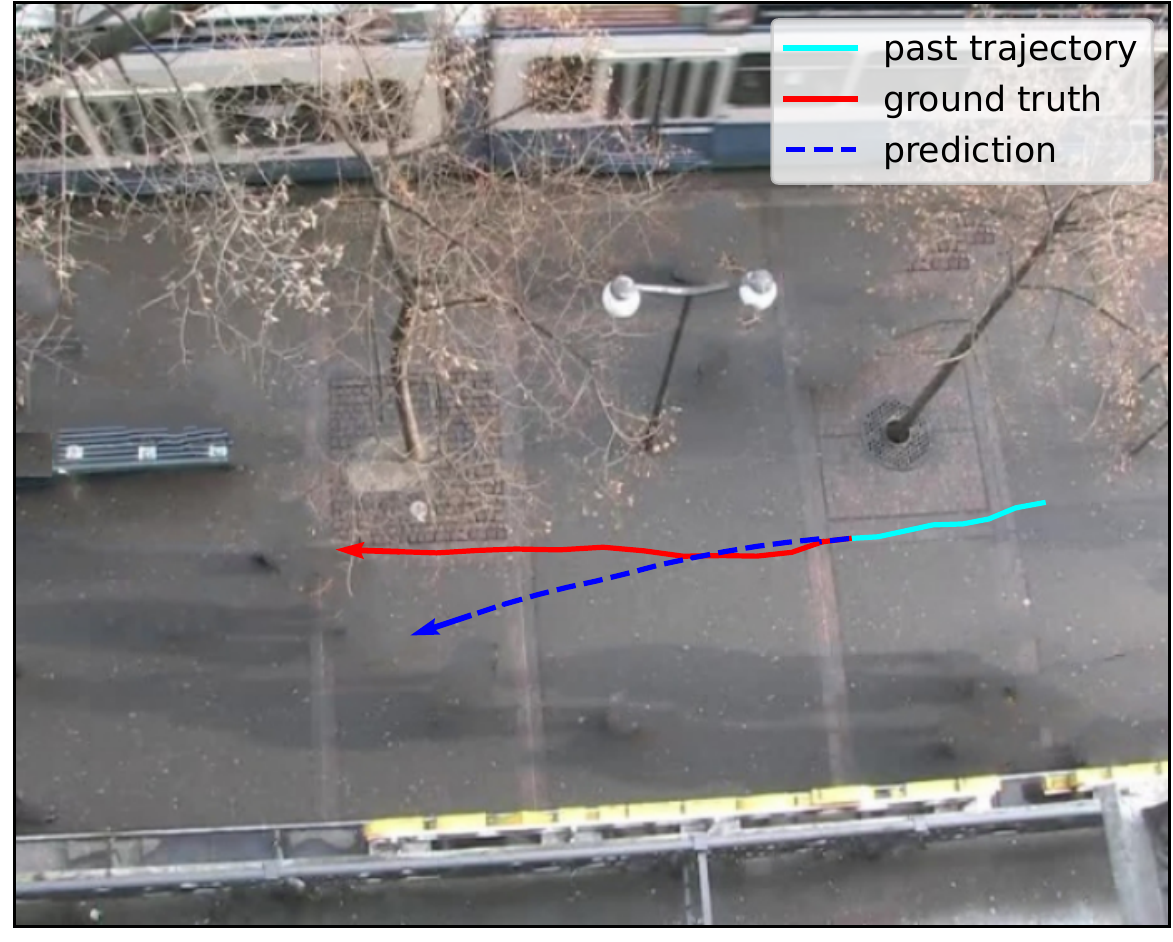}
     }
     \quad
     \subfigure{
         \includegraphics[trim=0in 0.1in 0.0in 0.0in, clip=true,width=0.3\linewidth]{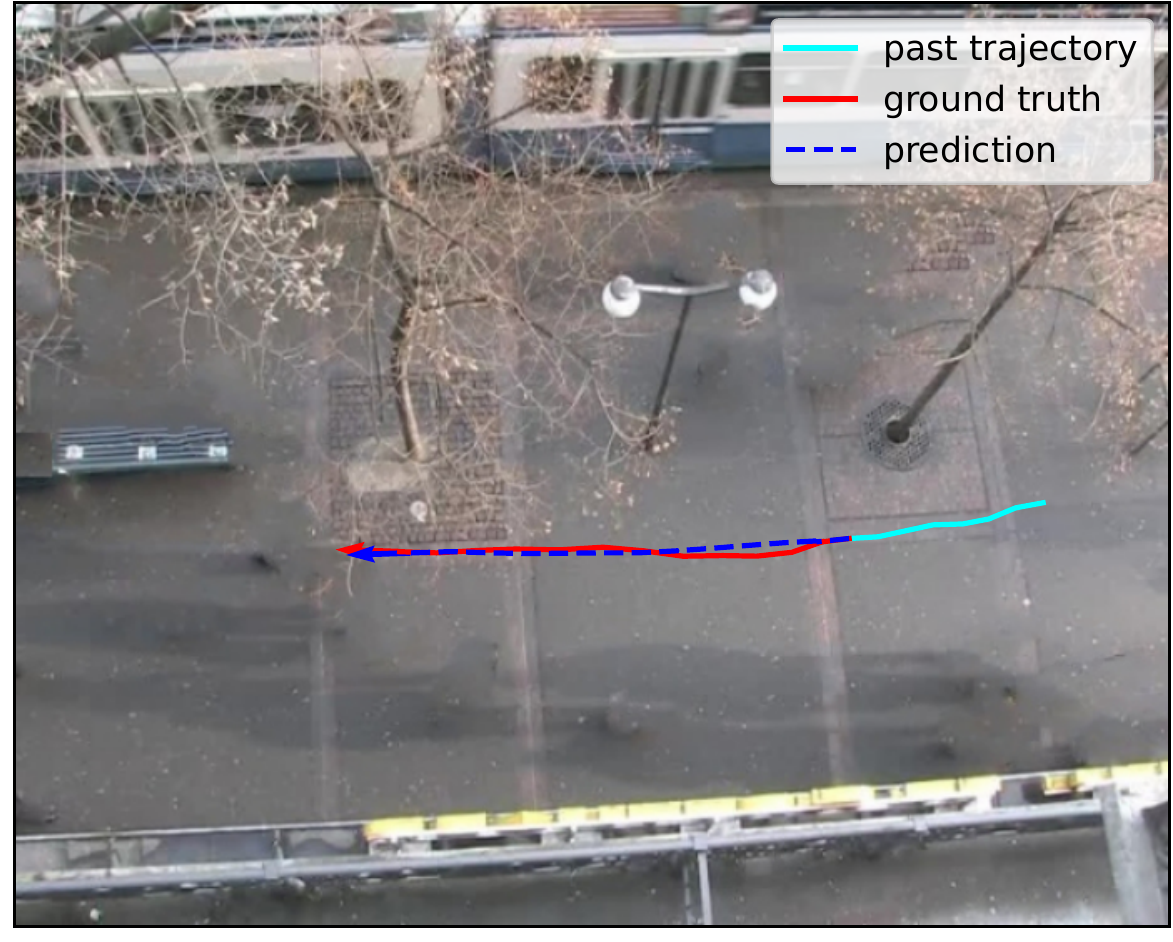}}
     \quad
     \subfigure{
         \includegraphics[trim=0in 0.1in 0.0in 0.0in, clip=true,width=0.3\linewidth]{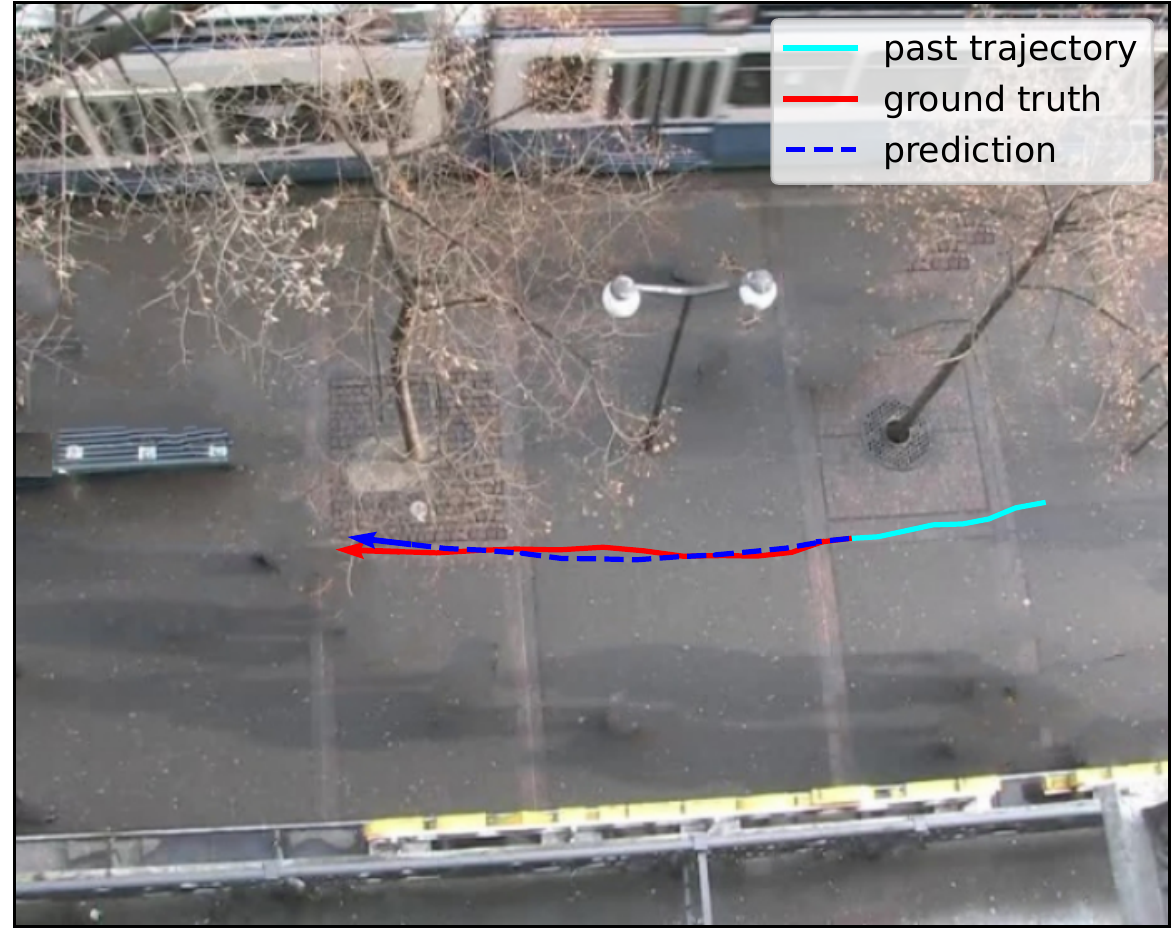}}
\caption{Prediction results generated by AgentFormer (left column), ForceFormer-Dr (middle column), and ForceFormer-Re (right column). Each row represents a different scene.} 
\label{fig:qualitative}
\end{figure*}
Figure \ref{fig:qualitative} shows the qualitative results predicted by AgentFormer (left column), ForceFormer-Dr (middle column), and ForceFormer-Re (right column), respectively.
From the upper row, we can see that, compared to AgentFormer, ForceFormer-Dr benefits from the goal information and driving force to predict trajectories around corners or turns. Although ForceFormer-Re predicts less accurate curving trajectories, its prediction for other trajectories is closer to the corresponding ground truth.
In the middle row for a scenario with two pedestrians walking in parallel, both ForceFormer-Dr and ForceFormer-Re predict more accurate final positions as the pedestrians make a left turn. In contrast, AgnetFormer does not explore the goal information from the goal-estimation module and predicts walking in the middle of the road.
A more visible scenario of predicting the final position can be seen in the bottom row. The prediction from AgentFormer largely deviates from the ground truth trajectory, while the predictions from ForceFormer-Dr and ForceFormer-Re are well aligned with the ground truth trajectory.

\vspace{6pt}
\noindent\textbf{Limitations.} 
Despite the enhanced performance brought by the social forces and the goal-estimation module, several limitations of the proposed model need to be noted.
The collisions have been reduced, but the predictions from ForceFormer are not totally collision-free.
One possible reason might be that in the social force module, we do not consider the interactions and forces within groups, which also may cause collisions. In future work, we will build a more comprehensive social force module and apply it to better simulate interactions among group members.
Moreover, the overall performance of ForceFormer, especially the calculation of social forces, relies on the reliability of the goal-estimation module.
Sub-optimal performance of this module can lead to compound errors in the final prediction. 
On the other hand, if we can access the ground truth goal information, we can quickly turn our model into a motion planning model.

\section{Conclusion}
This paper proposes a new goal-based trajectory predictor called ForceFormer that incorporates social forces into a Transformer-based generative model backbone. 
A U-Net-based goal-estimation module is adopted to predict the goals of pedestrians' trajectories. 
Additional to the position and velocity information, we derive the driving force from the estimated goal to efficiently simulate the guidance of a target on a pedestrian.
Also, repulsive forces are used to help the model learn collision avoidance among neighboring pedestrians.
ForceFormer achieves performance on par with the state-of-art models and better performance in the high-density scenarios on widely used pedestrian datasets.



\newpage
\bibliographystyle{IEEEtran}
\bibliography{mybib}

\end{document}